\documentclass[10pt,twocolumn,letterpaper]{article}

\usepackage{iccv}
\usepackage{times}
\usepackage{epsfig}
\usepackage{graphicx}
\usepackage{amsmath}
\usepackage{amssymb}

\usepackage{booktabs}
\usepackage{tikz}
\usepackage{comment}
\usepackage{stmaryrd}
\usepackage{multirow}
\usepackage{tabularx}
\usepackage{upquote}
\usepackage{bbm}
\usepackage{adjustbox}
\usepackage{physics}
\usepackage{pifont}
\usepackage{newfile}
\usepackage{xr}
\usepackage{caption}
\usepackage{mwe}
\usepackage{color, colortbl}
\usepackage{microtype}
\usepackage{dsfont}
\usepackage{lipsum}

\usepackage{makecell}

\newcommand{\cmark}{\ding{51}}

\definecolor{Gray}{gray}{0.9}

\usepackage[pagebackref=true,breaklinks=true,letterpaper=true,colorlinks,bookmarks=false]{hyperref}

\usepackage[capitalize]{cleveref}
\crefname{section}{Sec.}{Secs.}
\Crefname{section}{Section}{Sections}
\Crefname{table}{Table}{Tables}
\crefname{table}{Tab.}{Tabs.}

\iccvfinalcopy 

\def\iccvPaperID{6118} 

\ificcvfinal\fi

\begin{document}

\title{Knowing Where to Focus: Event-aware Transformer for Video Grounding}

\author{Jinhyun Jang$^1$ \quad\quad Jungin Park$^1$ \quad\quad Jin Kim$^1$ \quad\quad Hyeongjun Kwon$^1$ \quad\quad Kwanghoon Sohn$^{1,2}$\thanks{Corresponding author}\\
$^1$Yonsei University \quad\quad $^2$Korea Institute of Science and Technology (KIST)\\
{\tt\small $\lbrace$jr000192, newrun, kimjin928, kwonjunn01, khsohn$\rbrace$@yonsei.ac.kr}
}

\maketitle

\ificcvfinal\thispagestyle{empty}\fi
\let\thefootnote\relax\footnotetext{This research was supported by the National Research Foundation of Korea (NRF) grant funded by the Korea government (MSIP) (NRF2021R1A2C2006703).}

\begin{abstract}
\vspace{-7pt}
Recent DETR-based video grounding models have made the model directly predict moment timestamps without any hand-crafted components, such as a pre-defined proposal or non-maximum suppression, by learning moment queries.
However, their input-agnostic moment queries inevitably overlook an intrinsic temporal structure of a video, providing limited positional information.
In this paper, we formulate an event-aware dynamic moment query to enable the model to take the input-specific content and positional information of the video into account.
To this end, we present two levels of reasoning: 1) Event reasoning that captures distinctive event units constituting a given video using a slot attention mechanism; and 2) moment reasoning that fuses the moment queries with a given sentence through a gated fusion transformer layer and learns interactions between the moment queries and video-sentence representations to predict moment timestamps.
Extensive experiments demonstrate the effectiveness and efficiency of the event-aware dynamic moment queries, outperforming state-of-the-art approaches on several video grounding benchmarks.
The code is publicly available at \url{https://github.com/jinhyunj/EaTR}.
\end{abstract}

\vspace{-10pt}
\section{Introduction}
\label{sec:intro}

Over the decade, online video platforms have been explosively developed, with the number of videos uploaded every day growing exponentially.
Accordingly, the amount of work for video search (\eg video summarization~\cite{park2020sumgraph,jiang2022joint}, video retrieval~\cite{park2022probabilistic,wray2021semantic}, text-to-video retrieval~\cite{chen2021learning,patrick2020support}) has been explored to enable users to efficiently browse the information they want.
While they have presented an efficient way to search videos by considering the whole content, providing a user-defined moment in a video is a different desire.
As an alternative to this way, video grounding~\cite{anne2017localizing,liu2021adaptive,liu2022umt,moon2023query} has been explored in recent years.

\begin{figure}[!t]
    \centering
    \hfill
    \begin{subfigure}{1\linewidth}
    \centering
        \includegraphics[width=1\linewidth]{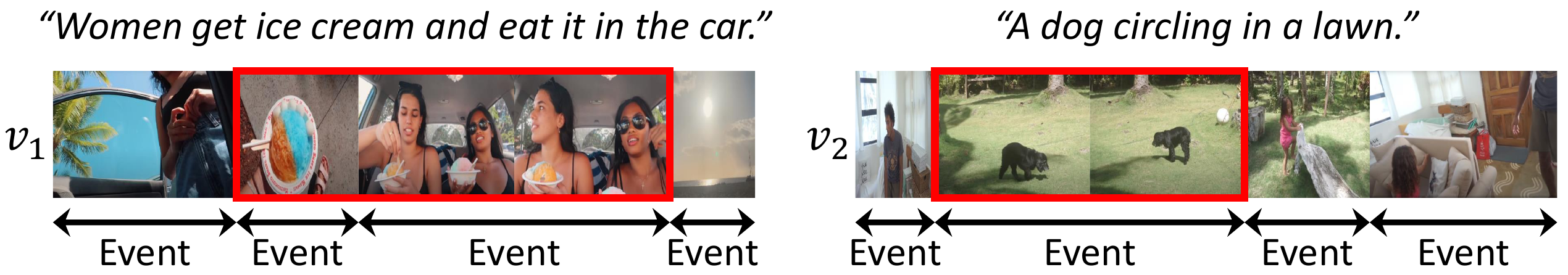}
        \vspace{-10pt}
        \caption{Video grounding}
        \label{fig:fig1a}
    \end{subfigure}
    \\
    \centering
    \hfill
    \begin{subfigure}{1\linewidth}
    \centering
        \includegraphics[width=1\linewidth]{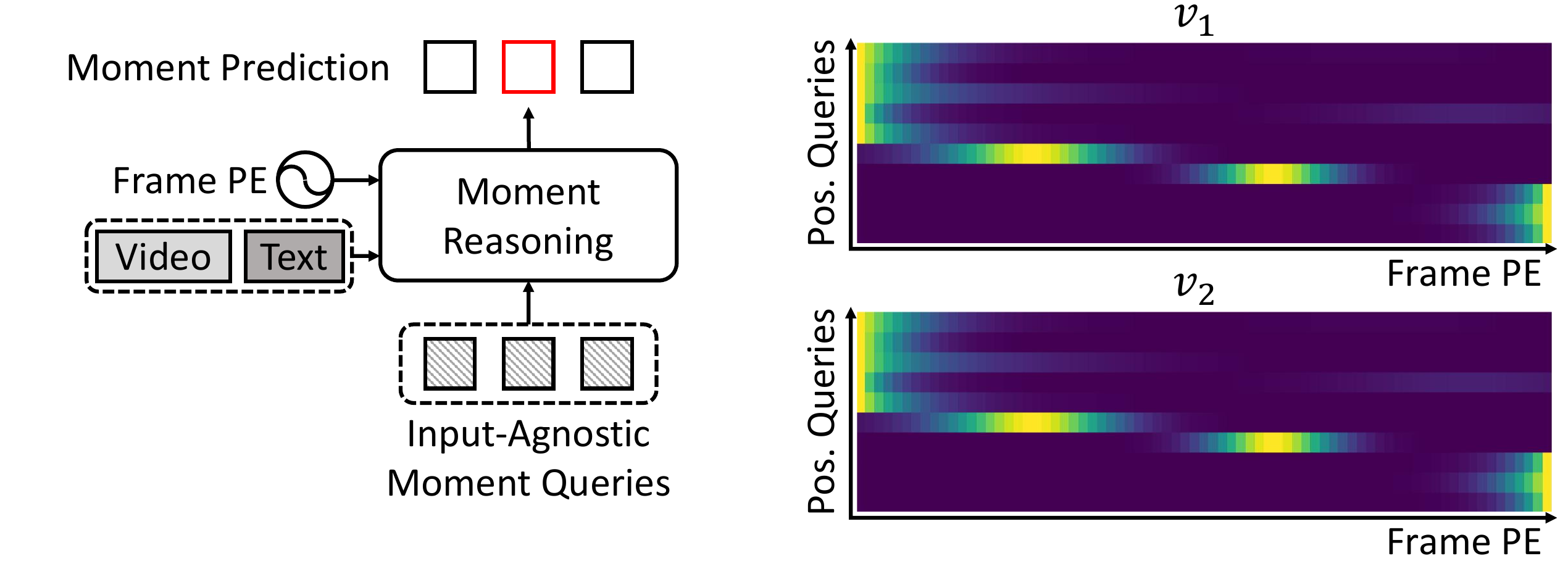}
        \caption{Previous DETR-based approach}
        \label{fig:fig1b}
    \end{subfigure}
    \\
    \centering
    \hfill
    \begin{subfigure}{1\linewidth}
    \centering
        \includegraphics[width=1\linewidth]{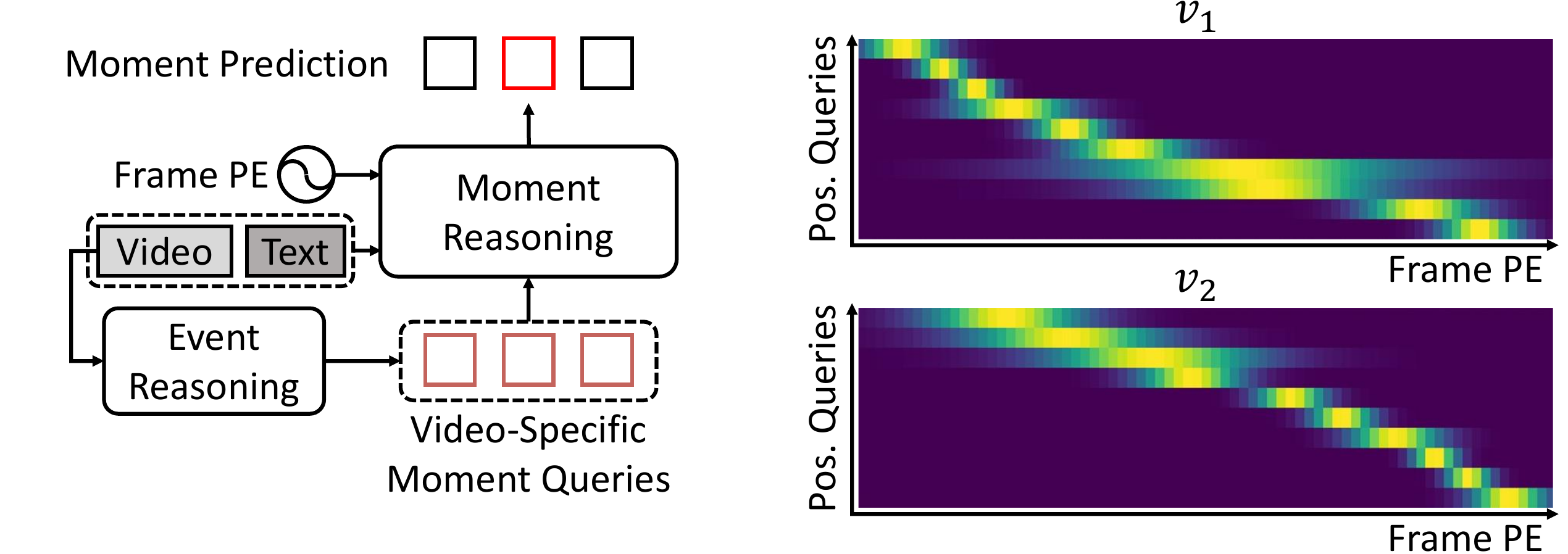}
        \caption{Ours}
        \label{fig:fig1c}
    \end{subfigure}
    \\
    \caption{(a) Video grounding aims to localize timestamps of a moment referring to a given sentence. Each video is composed of its own set of event units with varying lengths. 
    (b) Previous DETR-based methods learn input-agnostic moment queries, providing fixed referential search areas.
    (c) The proposed method learns event-aware moment queries, providing reliable referential search areas according to the given video.}
    \label{fig:fig1}
    \vspace{-10pt}
\end{figure}

Video grounding (also called natural language video moment localization) aims to localize timestamps of a moment referring to a given natural language sentence in a video, {as shown in \figref{fig:fig1a}}.
A key to identifying sentence-relevant moments is to 1) align video-language information; and 2) precisely reason temporal area.
Most works have accomplished this by aligning sentence and heuristically pre-defined temporal proposals (\eg sliding windows~\cite{anne2017localizing,gao2017tall,liu2018attentive}, temporal anchors~\cite{chen2018temporally,yuan2019semantic,zhang2019cross}) or directly learning sentence-frame interactions~\cite{lu2019debug,chen2020rethinking,zeng2020dense}.
However, they {highly} rely on the quality of hand-crafted components (\eg proposals, non-maximum suppression) to achieve a promising result.

The recent success of detection transformer (DETR)~\cite{carion2020end} has inspired approaches to integrate transformers~\cite{vaswani2017attention} into video grounding framework~\cite{lei2021detecting,cao2021pursuit,woo2022explore,fang2022hierarchical}.
{They learn a referential search area with a set of trainable embeddings, called \textit{moment queries}, as an alternative to the heuristically designed proposals.}
    {Each moment query probes and aggregates video-sentence representations through the cross-attention mechanism where the final moment queries are used to predict the timestamps of the sentence-relevant moments.}
    While they have achieved outperforming performance over the CNN-based approaches, the design choices of moment queries are still underexplored, {exhibiting several drawbacks.}
    Specifically, since moment queries are learned to contain general positional information, they produce a fixed input-agnostic search area during inference, as shown in \figref{fig:fig1b}.
    However, a video is a complex visual stream that consists of multiple semantic units (\ie, events) with varying lengths~\cite{hussein2019timeception,shou2021generic,sadhu2021visual,kang2022uboco}.
    In addition, the moment queries are controlled with equal contributions to aggregate video-sentence representations in the decoder layers, missing salient information and resulting in slow convergence.

        In this paper, we propose a novel \underline{E}vent-\underline{a}ware Video Grounding \underline{TR}ansformer (EaTR) that formulates a video as a set of event units and {treats video-specific event units as dynamic moment queries.}
    Our EaTR performs two different levels of reasoning: 1) Event reasoning that identifies the event units comprising the video and produces the content and positional queries; and 2) Moment reasoning that fuses the moment queries with the given sentence, and interacts with the video-sentence representations to predict the final timestamps for video grounding.
    Specifically, the randomly initialized learnable event slots identify the distinctive event units from the given video using a slot attention mechanism~\cite{locatello2020object}.
    The identified event units are then used as the moment queries in moment-level reasoning.
    While the moment queries in EaTR provide the input-specific referential search area as shown in \figref{fig:fig1c}, the interaction between the video-sentence representations and the sentence-relevant moment queries should be properly captured to predict an accurate moment timestamps.
    To this end, we introduce a gated fusion transformer layer to effectively minimize the impact of the sentence-irrelevant moment queries and capture the most informative referential search area.
    In the gated fusion transformer layer, we fuse the moment queries and sentence representation according to their similarity in order to adaptively aggregate sentence information to the informative moment queries.
    The fused moment queries interact with the video-sentence representations in the transformer decoder to make the final decision for video grounding.

    {Extensive experiments on several video grounding benchmarks~\cite{lei2021detecting,gao2017tall,krishna2017dense} demonstrate the effectiveness of the event-aware video grounding framework, achieving a new state-of-the-art performance over the previous methods~\cite{lei2021detecting,liu2022umt,zhu2023rethinking}.
    In summary, our key contributions are as follows:
    (i) We present a novel Event-aware Video Grounding Transformer (EaTR) that enhances the temporal reasoning capability of the moment queries by learning the video-specific event information.
    (ii) We introduce effective event reasoning and the gated fusion that highlight the distinctive events in a given video and sentence.
    (iii) We conduct extensive experiments to validate the effectiveness of the proposed method, and outperform state-of-the-art approaches on three video grounding benchmarks, including QVHighlights~\cite{lei2021detecting}, Charades-STA~\cite{gao2017tall}, and ActivityNet Captions~\cite{krishna2017dense}.}

\section{Related Work}
\label{sec:related}

\paragraph{Video grounding.}

A standard framework of localizing a moment corresponding to a given sentence can be categorized into two different paradigms.
(i) Proposal-driven approaches first generate several candidate proposals and rank them based on their similarity with a sentence.
Most works utilize pre-defined proposals such as sliding windows~\cite{anne2017localizing,gao2017tall,liu2018attentive,liu2018cross,ge2019mac,zhang2019exploiting} or temporal anchors~\cite{chen2018temporally,yuan2019semantic,zhang2019cross,zhang2019man,liu2020jointly}.
Others proposed to generate high-quality proposals by exploring every possible pairs of start-end points~\cite{zhang2020learning,liu2021context,xiao2021boundary} or with sentence guidance~\cite{shao2018find,xu2019multilevel,chen2019semantic,liu2021adaptive}.
(ii)
Proposal-free approaches directly predict the target moment via learning video-sentence interactions.
Several works attempt to solve the problem by formulating attention mechanisms~\cite{yuan2019find,ghosh2019excl,zhang2020span,rodriguez2020proposal,mun2020local}, making dense predictions~\cite{lu2019debug,chen2020rethinking,zeng2020dense}, combining complementary visual features (\eg object regions, motion features)~\cite{zeng2021multi,chen2021end,liu2022exploring}, and reducing the dataset bias~\cite{otani2020uncovering,zhang2021towards,yang2021deconfounded,nan2021interventional,liu2022memory,hao2022can}.
Although the two paradigms have achieved impressive results, they are limited in their use of hand-crafted components (\eg pre-defined proposals, non-maximum suppression) and redundancy (\eg large number of candidate proposals).

To simplify the whole process into an end-to-end manner, recent works~\cite{lei2021detecting,cao2021pursuit,woo2022explore,liu2022umt,fang2022hierarchical,xu2023mh,moon2023query} adopted DETR-based architecture into the video grounding task.
Despite the progress, the ineffective use of learnable moment queries limits the model capability on temporal reasoning.
UMT~\cite{liu2022umt} attempted to improve the query design by conditioning them on extra modality (\eg audio, optical flow).
Instead of relying on additional input, we propose to utilize a video itself as an interpretable positional guidance.

\begin{figure*}[!t]
    \centering
    \includegraphics[width=0.99\linewidth]{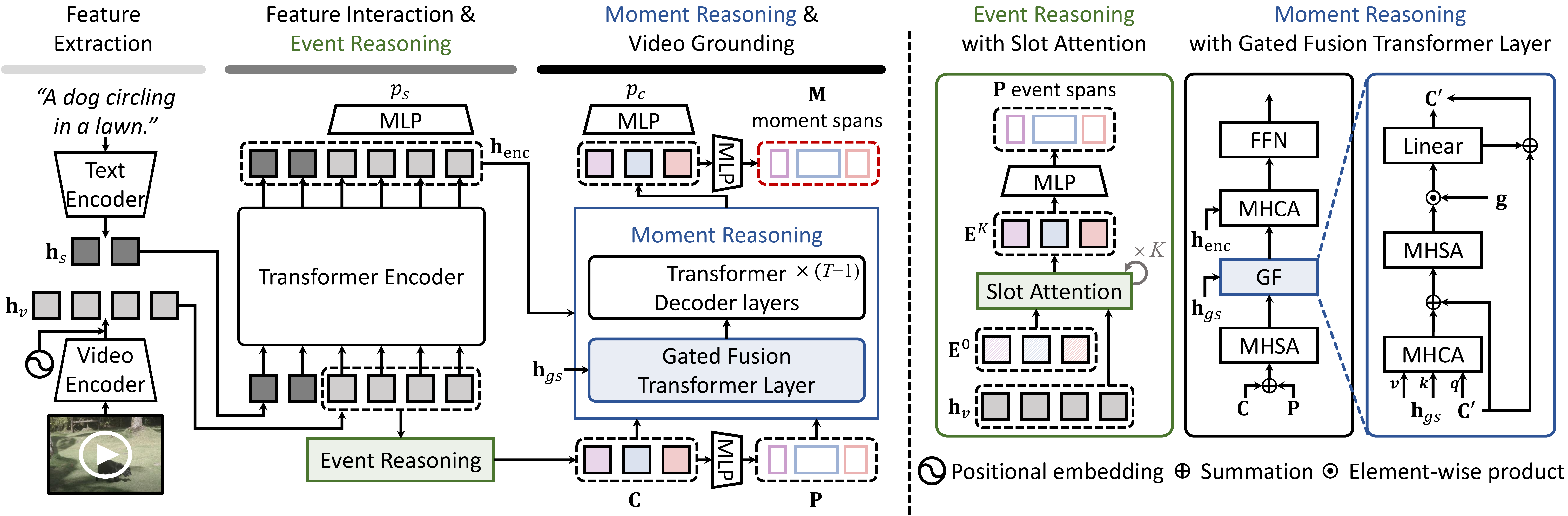}\vspace{5pt}
    \caption{An overview of the proposed EaTR.
    The procedure consists of three steps: 1) feature extraction, 2) feature interaction and event reasoning, and 3) moment reasoning for video grounding.
    The event reasoning is done through a slot attention mechanism with learnable event slots ($\mathbf{E}^0$).
    The outputs of the event reasoning network ($\mathbf{E}^K, {\mathbf{P}}$) serve as an initial moment queries ($\mathbf{C}, \mathbf{P}$) of the moment reasoning network.
    The initial moment queries are fused with the sentence feature ($\mathbf{h}_{gs}$) through a gated fusion transformer layer in order to better focus on the sentence-relevant queries.
    The queries then interact with the video-sentence representations (${\mathbf{h}_\text{enc}}$) for making the final prediction on the moment timestamps (${\mathbf{M}}$).
    }
    \label{fig:fig2}
\end{figure*}

\vspace{-5pt}
\paragraph{DETR and its variants.}
The adoption of transformers~\cite{vaswani2017attention} to object detection (DETR)~\cite{carion2020end} has streamlined the whole pipeline by removing the need of hand-crafted components while improving the performance.
Despite its success, DETR has its own issue of slow training convergence.
Several studies~\cite{zhu2020deformable,sun2021rethinking,yao2021efficient,gao2021fastconvergence,meng2021conditional,dai2021dynamic,wang2022anchor,zhang2022accelerating,liu2022dab,li2022dn} attribute the issue to the naive design of object query and its operation on cross-attention module;
object queries require a long training time to accurately learn where and what to focus at the cross-attention module.
The seminal work discovered the importance of spatial priors for convergence speed and performance by reformulating object queries as 2D center coordinates (\ie, \textit{x,y})~\cite{meng2021conditional,wang2022anchor} or 4D box coordinates (\ie, \textit{x,y,w,h})~\cite{zhu2020deformable,gao2021fastconvergence,liu2022dab,li2022dn}.
Others imposed spatial-constraint at the cross-attention module such as sparse sampling~\cite{zhu2020deformable}, Gaussian prior~\cite{gao2021fastconvergence}, or conditional weight~\cite{meng2021conditional,liu2022dab}.

\vspace{-5pt}
\paragraph{Generic event boundary detection.}
Generic Event Boundary Detection (GEBD)~\cite{shou2021generic} is a recently introduced video understanding task that aims to identify every instant that human perceive as event boundaries.
The event boundary includes a change of subject, action, and environment.
Recent works~\cite{kang2022uboco,tang2022progressive} explored frame-wise similarity, namely Temporal Self-similarity Matrix (TSM), to better perceive temporal variations.
Specifically, UBoCo~\cite{kang2022uboco} focused on producing boundary-sensitive features (\ie, distinctive TSM) in an unsupervised manner with a recursive TSM parsing mechanism.
Having a common goal of identifying the events without supervision, we adopt their contrastive kernel for generating pseudo event information.
However, instead of directly adopting the whole process, we newly design the event reasoning network suitable for the DETR-based video grounding network.

\vspace{-5pt}
\paragraph{Slot attention.}
Slot attention~\cite{locatello2020object} is a recently proposed iterative attention mechanism that aims to learn the object-centric representation.
The randomly initialized \textit{slots} are introduced to interact with the input features and group the pixels belonging to the same object.
Our event reasoning network employs the grouping property of the slot attention mechanism to aggregate visually similar frames into multiple events and train the network with the event localization loss.

\section{Proposed Method}\label{sec:method}

\subsection{Background and motivation}
Video grounding aims to localize the timestamps of moments referring to a given sentence $\mathcal{S}$ in an untrimmed video $\mathcal{V}$.
    {Recent DETR-based methods~\cite{lei2021detecting,cao2021pursuit,xu2023mh,moon2023query} formulate learnable query embeddings $\mathbf{Q}$ (\ie, moment query) that represent a set of learnable referential search areas, and predict the target moments using transformer~\cite{vaswani2017attention} in an end-to-end manner.
    The moment query can be decomposed into two parts according to their physical roles: a content query $\mathbf{C}$ and a positional query $\mathbf{P}$.
    Each query is responsible for aggregating video-sentence representations based on a semantic similarity (with the content query) and a positional similarity (with the positional query).
    The previous works define the initial content and positional queries as zero embeddings and learnable embeddings respectively, and progressively aggregate video-sentence information by going through the transformer decoder.
    While they have successfully demonstrated the effectiveness of the DETR-based architecture, the design of the input-agnostic moment query makes the search area ambiguous and training difficult.}

  {To address this problem, we propose an event-aware video grounding transformer (EaTR) where a video is treated as a set of event units.
    In our framework, we formulate the dynamic moment queries to provide precise referential search area by identifying the event units from the given video, and fuse the dynamic moment queries with sentence information.
    In the following section, we describe feature extraction (\secref{sec:feature}), event reasoning (\secref{sec:event_reasoning}), and moment reasoning (\secref{sec:moment_reasoning}).
    The overall architecture is shown in \figref{fig:fig2}.}

\subsection{Feature extraction and interaction}\label{sec:feature}
    {Given a video $\mathcal{V}$ of length $L_v$ and a sentence $\mathcal{S}$ of length $L_s$, we first encode the video and sentence representations using corresponding pretrained backbone networks (\eg I3D~\cite{carreira2017quo}, CLIP~\cite{radford2021learning}) as follows:
    \begin{equation}\small
        \mathbf{h}_v = f_v(\mathcal{V}) + PE \in \mathbb{R}^{L_v \times d}, \quad \mathbf{h}_s = f_s(\mathcal{S}) \in \mathbb{R}^{L_v \times d},
    \end{equation}
    where $\mathbf{h}_v$ and $\mathbf{h}_s$ are the video and sentence representations, $PE$ is a set of positional embeddings for video frames, $f_v(\cdot)$ and $f_s(\cdot)$ are the pretrained backbone networks for each modality, respectively.
    Similar to Moment-DETR~\cite{lei2021detecting}, the video and sentence representations interact with each other through a stack of $T$ transformer encoder layers to obtain the video-sentence representations $\mathbf{h}_\text{enc}$:
    \begin{equation}
        \mathbf{h}_\text{enc} = f_\text{enc}(\mathbf{h}_v || \mathbf{h}_s),
    \end{equation}
    where $f_\text{enc}(\cdot)$ is the transformer encoder and $||$ is the concatenate operation.
    Following \cite{lei2021detecting}, we feed the partial representations corresponding to the video in $\mathbf{h}_\text{enc}$ into a linear layer to predict the saliency score $p_s\in \mathbb{R}^{L_v}$ that represents the similarity between the video frames (or clip) and words in the sentence.
    To enlarge the saliency score gap between the sentence-relevant frames and other frames, we employ the saliency loss~\cite{lei2021detecting}:
    \begin{equation}
        \mathcal{L}_\text{sal}=\text{max}(0, \alpha + \bar{p}_{s,\text{out}} - \bar{p}_{s,\text{in}}),
    \end{equation}
    where $\bar{p}_{s,\text{in}}$ and $\bar{p}_{s,\text{out}}$ are the average saliency scores of {randomly sampled} frames within and outside of the ground truth time interval, respectively, and $\alpha$ is a margin.
    }

    \subsection{Event reasoning}\label{sec:event_reasoning}
    {The main problem of the previous moment queries is that they provide an ambiguous referential search area due to the design of the input-agnostic moment queries.
    To handle the problem, we propose to identify the distinctive event units from the given video and utilize event information for initializing the dynamic moment queries.}

    {Specifically, we derive $N$ event units from the video representation $\textbf{h}_v$ using a set of $N$ learnable event slots $\mathbf{E}\in \mathbb{R}^{N \times d}$ and the slot attention mechanism~\cite{locatello2020object}.
    Formally, the event slots iteratively interact with the video representations $\mathbf{h}_v$ for $K$ iterations to group the visually similar frames and obtain the final event units $\mathbf{E}^K$.
    In $k$-th iteration, we first embed $\mathbf{h}_v$ and $\mathbf{E}^k$ using layer normalization followed by linear projections, such that:
    \begin{equation}
        \mathbf{h}'_v = (\texttt{LN}(\mathbf{h}_v))\mathbf{W}_1, \quad \mathbf{E}'^{k-1} = (\texttt{LN}(\mathbf{E}^{k-1}))\mathbf{W}_2,
    \end{equation}
    where $\mathbf{h}'_v$ and $\mathbf{E}'^{k-1}$ are the embedded video representations and event slots, $\texttt{LN}(\cdot)$ is layer normalization, $\mathbf{W}_1$, and $\mathbf{W}_2$ are the linear projection matrices.
    The $k$-th interaction matrix between $\mathbf{h}'_v$ and $\mathbf{E}'^{k-1}$ can be computed as:
    \begin{equation}
        \mathbf{A}^k = \texttt{Softmax}\left(\frac{(\mathbf{h}'_v)(\mathbf{E}'^{k-1})^\top}{\sqrt{d}}\right) \in \mathbb{R}^{L_v \times N},
    \end{equation}
    where $\texttt{Softmax}()$ is the softmax function along the event slot direction.
    With the interaction matrix $\mathbf{A}^k$, the $k$-th event slots are updated by following equation:
    \begin{equation}\small
        \begin{aligned}
            \mathbf{U} = (\hat{\mathbf{A}}^k)^\top (\mathbf{h}_v)&\mathbf{W}_3\;+\; \mathbf{E}^{k-1}, \;\; \text{where} \;\; \hat{\mathbf{A}}^k_{l,n}=\frac{\mathbf{A}^k_{l,n}}{\sum_{L_v}\mathbf{A}^k_{l,n}},\\
                &\mathbf{E}^k=(\texttt{LN}(\mathbf{U}))\mathbf{W}_4 + \mathbf{U},
        \end{aligned}
    \end{equation}
    with additional linear projection matrices $\mathbf{W}_3$ and $\mathbf{W}_4$.
    Different from the conventional slot attention~\cite{locatello2020object}, we replace the GRU layer with residual summation to avoid inefficient computations.
    Since each event slot $\mathbf{e}_n^K \in \mathbf{E}^K$ contains visual information corresponding to the individual event of the video, we use $\mathbf{E}^K$ as the initial content queries $\mathbf{C}$.
    In addition, we project $\mathbf{E}^K$ to the 2-dimensional embedding space to derive the initial positional queries $\mathbf{P}$ that represent the center and duration of the referential search area for each moment query as:
    \begin{equation}
        \mathbf{P} = \mathbf{E}^K\mathbf{W}_p = \lbrace (c_n, w_n) \rbrace_{n=1}^N,
    \end{equation}
    where $c_n$ and $w_n$ are the center and width of the referential time span for the $n$-th content query.
    While cross-attention can be used as an alternative to slot attention, slot attention achieves superior performance with higher efficiency by enforcing the slots to compete each other and reusing linear projection matrices (\eg $\mathbf{W}_1$) for every iteration.

    {To guarantee the moment queries contain the event units, we learn event reasoning by generating the pseudo event timestamps of the video based on the temporal self-similarity matrix~\cite{panagiotakis2018unsupervised,dwibedi2020counting,nam2021zero,kang2022uboco}.
    Specifically, we employ the recent contrastive kernel~\cite{kang2022uboco} that computes the event boundary scores by convolving with the diagonal elements of TSM.
    By thresholding and sampling the boundary scores, we are able to obtain the timestamps of the event units $\hat{\mathbf{P}}$, where each element $\hat{\mathbf{P}}_i\in[0,1]^2$ defines the normalized center coordinate and duration of an event.
    With the pseudo event timestamps, we formulate an event localization loss between positional queries and each corresponding pseudo event timestamp.
    Since the order of the predicted event sets is arbitrary, we find an optimal assignment between the positional queries and pseudo event spans via the Hungarian matching algorithm~\cite{kuhn1955hungarian,carion2020end}.} 
The optimal assignment $\hat{\sigma}$ is determined based on the similarity of the pseudo event spans $\hat{\mathbf{P}}$ and predicted event spans ${\mathbf{P}}$ as:
\begin{equation}\label{eq:optimal_event}\small
\begin{gathered}
    \hat{\sigma}=\underset{\sigma\in\mathfrak{G}_N}{\text{arg min}}\sum\limits_i^N\mathcal{C}(\hat{\mathbf{P}}_i,{\mathbf{P}}_{\sigma(i)}),\\
    \mathcal{C}(\hat{\mathbf{P}}_i,{\mathbf{P}}_{\sigma(i)})=\lambda_{l_1}||\hat{\mathbf{P}}_i-{\mathbf{P}}_{\sigma(i)}||_1+\lambda_\text{iou}\mathcal{L}_\text{iou}(\hat{\mathbf{P}}_i,{\mathbf{P}}_{\sigma(i)}),
\end{gathered}
\end{equation}
where $\mathcal{L}_\text{iou}$ is generalized temporal IoU~\cite{rezatofighi2019generalized}.
$\lambda_{l_1}$ and $\lambda_\text{iou}$ are the balancing parameters.
Given the optimal assignment, the event localization loss is defined as:
\begin{equation}\label{eq:event}
\begin{aligned}
    \mathcal{L}_\text{event}=\sum\limits_i^N\mathcal{C}(\hat{\mathbf{P}}_i,{\mathbf{P}}_{\hat{\sigma}(i)}).
\end{aligned}
\end{equation}

\subsection{Moment reasoning}
\label{sec:moment_reasoning}
    {In moment reasoning, we aggregate the video-sentence representations to the moment queries through a stack of $T$ transformer decoder layers to predict the final moment timestamps.
    Although the initial dynamic moment queries contain the video-specific referential search areas, the sentence-relevant moment queries should be enhanced while filtering out irrelevant queries to produce more reliable search areas.
    To this end, we propose a gated fusion (GF) transformer layer that enhances the sentence-relevant moment queries and suppresses the other queries by fusing the moment queries with a global sentence representation.
    Our GF transformer layer is heavily inspired by \cite{wang2019camp,liu2021progressively}, but transformed into a form suitable to the transformer architecture.
    
    \vspace{-5pt}
    \paragraph{Enhanced moment query.}
    Each positional query $\mathbf{p}_n = (c_n, w_n)$ is first extended to $d$-dimensional space through the sinusoidal positional encoding (PE)~\cite{meng2021conditional,liu2022dab}, concatenation, and MLP layer as:
    \begin{equation}
        \mathbf{p}_n \leftarrow \texttt{MLP}(\texttt{Concat}(PE(c_n), PE(w_n))).
    \end{equation}
    By expanding the positional queries to $d$-dimensional space, the multi-head self-attention (MHSA) and cross-attention (MHCA) layers can take the content and positional queries as inputs at the same level.
    Before fusing the moment queries and global sentence representation, we feed the sum of the content and positional queries into the MHSA layer, such that the enhanced moment query $\mathbf{C}'$ can be obtained by:
    \begin{equation}
        \mathbf{C}' = \texttt{MHSA}(\mathbf{C} \oplus \mathbf{P})\in \mathbb{R}^{N \times d},
    \end{equation}
    where $\oplus$ is an element-wise summation.

    \vspace{-5pt}
    \paragraph{Gated fusion (GF) transformer layer.}
    The GF transformer layer takes $\mathbf{C}'$ and the global sentence representation $\mathbf{h}_{gs}$ which is obtained by applying max pooling on $\mathbf{h}_s$.
    We treat $\mathbf{C}'$ as the query and $\mathbf{h}_{gs}$ as the key and value of the MHCA layer to aggregate the query-relevant sentence information, such that the aggregated sentence representations $\hat{\mathbf{C}}$ is derived as:
    \begin{equation}
        \hat{\mathbf{C}} = \texttt{MHCA}(\mathbf{C}', \mathbf{h}_{gs}, \mathbf{h}_{gs}) \in \mathbb{R}^{N \times d}.
    \end{equation}
    The similarity between the $n$-th moment query $\mathbf{c}'_n$ and aggregated sentence representation $\hat{\mathbf{c}}_n$ is then used as a gate to suppress the irrelevant queries.
    In other words, the gate for the sentence-relevant query has a high value, otherwise represents a low value.
    The gate for the $n$-th moment query is computed as a single scalar by:
    \begin{equation}\label{eq:gate}
        g_n = \texttt{Sigmoid}(\mathbf{c}'_n \cdot \hat{\mathbf{c}}_n^\top).
    \end{equation}
    The gated fusion is then formulated with $\mathbf{C}'$, $\hat{\mathbf{C}}$, and the gate $\mathbf{g}$ by the following equation:
    \begin{equation}\label{eq:gate_fuse}
        {\mathbf{C}'} \leftarrow \texttt{Linear}(\mathbf{g}\odot\texttt{MHSA}(\mathbf{C}'\oplus\hat{\mathbf{C}})) + \mathbf{C}',
    \end{equation}
    where $\texttt{Linear}$ is a single linear layer and $\odot$ is the element-wise multiplication.
    The enhanced moment queries then interact with the video-sentence representations $\mathbf{h}_\text{enc}$ through the {modulated MHCA~\cite{liu2022dab,moon2023query}} and are fed to a feed-forward network (FFN).

\vspace{-5pt}
    \paragraph{Moment prediction.}
    After the GF transformer layer, the moment queries go through the remaining $(T-1)$ transformer decoder layers {where each positional query is updated layer-wise with its offset predicted from the corresponding content query~\cite{zhu2020deformable,liu2022dab}.}
    The output of the transformer decoder $\mathbf{h}_\text{dec}$ is fed to FFN for predicting the moment span ${\mathbf{M}}$.
    We also utilize a linear layer to predict the confidence score $p_c\in\mathbb{R}^N$ corresponding to each moment query.
To learn the moment localization, set prediction loss based on bipartite matching is applied~\cite{carion2020end}.
Given the ground truth moment timestamps $\hat{\mathbf{M}}_i\in[0,1]^2$ consisting of the normalized center coordinate and width, the optimal assignment is determined based on the timestamp similarities and the corresponding confidence scores using the Hungarian algorithm as:
\begin{equation}\label{eq:optimal}\small
\begin{gathered}
    \hat{\sigma}'=\underset{\sigma'\in\mathfrak{G}_N}{\textrm{arg min}}\sum\limits_i^N\left[-\lambda_cp_{c,{\sigma'(i)}}+\mathcal{C}(\hat{\mathbf{M}}_i,{\mathbf{M}}_{\sigma'(i)})\right],\\
\begin{aligned}
    \mathcal{C}(\hat{\mathbf{M}}_i,{\mathbf{M}}_{\sigma'(i)})=&\lambda_{l_1}||\hat{\mathbf{M}}_i-{\mathbf{M}}_{\sigma'(i)}||_1\\
    &+\lambda_\textrm{iou}\mathcal{L}_\textrm{iou}(\hat{\mathbf{M}}_i,{\mathbf{M}}_{\sigma'(i)}),
\end{aligned}
\end{gathered}
\end{equation}
where $\lambda_{l_1}, \lambda_\text{iou}$  and $\lambda_c$ are the balancing parameters.
With the optimal assignment $\hat{\sigma}'$, the moment localization loss~\cite{cao2021pursuit,lei2021detecting} is defined as:
\begin{equation}\label{eq:moment}
\begin{aligned}
    \mathcal{L}_\textrm{moment}=\sum\limits_i^N\left[-\lambda_{c}\log p_{c,{\hat{\sigma}'(i)}}+\mathcal{C}(\hat{\mathbf{M}}_i,{\mathbf{M}}_{\hat{\sigma}'(i)})\right].
\end{aligned}
\end{equation}

\vspace{-5pt}
\paragraph{Overall objectives.}
The overall objective is defined as:
\begin{equation}\label{eq:overall}
\begin{aligned}
    \mathcal{L}_\text{overall}=\mathcal{L}_\text{moment}+\lambda_\text{sal}\mathcal{L}_\text{sal}+\lambda_\text{event}\mathcal{L}_\text{event},
\end{aligned}
\end{equation}
where $\lambda_\text{sal}$ and $\lambda_\text{event}$ are the balancing parameters.

\begin{table*}[!t]
\caption{Experimental results on QVHighlights val split. HD represents highlight detection. We repeat the experiment with 5 different seeds and report the mean performance and standard deviation. $^\dag$ indicates the model with additional audio input.}
\begin{center} \label{tab:qv}
\resizebox{0.85\linewidth}{!}{
\setlength{\tabcolsep}{4pt}
\begin{tabular}{lccccccccc}
    \toprule
    \multirow{3}{*}{{Methods}} & \multicolumn{5}{c}{{Video Grounding}} & \multicolumn{2}{c}{{HD}} & \multirow{3}{*}{GFLOPs} & \multirow{3}{*}{Params}\\
    \cmidrule(lr){2-6} \cmidrule(lr){7-8}
    & \multicolumn{2}{c}{R1} & \multicolumn{3}{c}{mAP} & \multicolumn{2}{c}{$\geq$ Very Good} \\
    \cmidrule(lr){2-3} \cmidrule(lr){4-6} \cmidrule(lr){7-8}
    & @0.5 & @0.7 & @0.5 & @0.75 & Avg. & mAP & HIT@1 \\
    \midrule
    BeautyThumb~\cite{song2016click} & - & - & - & - & - & 14.36 & 20.88 & - & - \\
    DVSE~\cite{liu2015multi} & - & - & - & - & - & 18.75 & 21.79  & - & - \\
    MCN~\cite{anne2017localizing} & 11.41 & 2.72 & 24.94 & 8.22 & 10.67 & - & -  & - & -  \\
    CAL~\cite{escorcia2019temporal} & 25.49 & 11.54 & 23.40 & 7.65 & 9.89 & - & -  & - & - \\
    CLIP~\cite{radford2021learning} & 16.88 & 5.19 & 18.11 & 7.00 & 7.67 & 31.30 & 61.04  & - & - \\
    XML~\cite{lei2020tvr} & 41.83 & 30.35 & 44.63 & 31.73 & 32.14 & 34.49 & 55.25 & - & - \\
    XML+~\cite{lei2020tvr} & 46.69 & 33.46 & 47.89 & 34.67 & 34.90 & 35.38 & 55.06  & - & - \\
    Moment-DETR~\cite{lei2021detecting} & 52.89$_{\pm2.3}$ & 33.02$_{\pm1.7}$ & 54.82$_{\pm1.7}$ & 29.40$_{\pm1.7}$ & 30.73$_{\pm1.4}$ & 35.69$_{\pm0.5}$ & 55.60$_{\pm1.6}$ & 0.28 & 4.8M \\
    UMT$^\dag$~\cite{liu2022umt} & 56.23 & 41.18 & 53.83 & 37.01 & 36.12 & {38.18} & {59.99} & 0.63 & 14.9M \\
    MH-DETR~\cite{xu2023mh} & 60.05 & 42.48 & 60.75 & 38.13 & 38.38 & 38.22 & 60.51 & 0.34 & 8.2M \\
    QD-DETR~\cite{moon2023query} & \textbf{62.40}$_{\pm1.1}$ & 44.98$_{\pm0.8}$ & \textbf{62.52}$_{\pm0.6}$ & 39.88$_{\pm0.7}$ & 39.86$_{\pm0.6}$ & \textbf{38.94}$_{\pm0.4}$ & \textbf{62.40}$_{\pm1.4}$ & 0.60 & 7.6M \\
    \midrule
    Ours & {61.36}$_{\pm1.2}$ & \textbf{45.79}$_{\pm0.7}$ & {61.86}$_{\pm0.6}$ & \textbf{41.91}$_{\pm0.6}$ & \textbf{41.74}$_{\pm0.7}$ & 37.15$_{\pm0.5}$ & 58.65$_{\pm1.4}$ & 0.47 & 9.0M \\
    \bottomrule
\end{tabular}}
\end{center}
\vspace{-10pt}
\end{table*}

\begin{table}[!t]
\caption{Experimental results on Charades-STA test split with I3D features and ActivityNet Captions val\_2 split with C3D features.}
\begin{center} \label{tab:ch_anet}
\resizebox{0.95\linewidth}{!}{
\setlength{\tabcolsep}{7pt}
\begin{tabular}{lcccc}
    \toprule
    \multirow{2}{*}{{Methods}} & \multicolumn{2}{c}{Charades-STA} & \multicolumn{2}{c}{ActivityNet Captions} \\
    \cmidrule(lr){2-3} \cmidrule(lr){4-5}
    & R1@0.5 & R1@0.7 & R1@0.5 & R1@0.7 \\
    \midrule
    BPNet~\cite{xiao2021boundary} & 50.75 & 31.64 & 42.07 & 24.69 \\
    DRN~\cite{zeng2020dense} & 53.09 & 31.75 & 45.45 & 24.36 \\
    FIAN~\cite{qu2020fine} & 58.55 & 37.72 & 47.90 & 29.81 \\
    LGI~\cite{mun2020local} & 59.46 & 35.48 & 41.51 & 23.07 \\
    DeNet~\cite{zhou2021embracing} & 59.70 & 38.52 & 43.79 & - \\
    CPN~\cite{zhao2021cascaded} & 59.77 & 36.67 & 45.10 & 28.10 \\
    CSMGAN~\cite{liu2020jointly} & 60.04 & 37.34 & 49.11 & 29.15 \\
    SSCS~\cite{ding2021support} & 60.75 & 36.19 & 46.67 & 27.56 \\
    CBLN~\cite{liu2021context} & 61.13 & 38.22 & 48.12 & 27.60 \\
    IA-Net~\cite{liu2021progressively} & 61.29 & 37.91 & 48.57 & 27.95 \\
    APGN~\cite{liu2021adaptive} & 62.58 & 38.86 & 48.92 & 28.64 \\
    MGSL-Net~\cite{liu2022memory} & 63.98 & 41.03 & 51.87 & 31.42 \\
    SMIN~\cite{wang2021structured} & 64.06 & 40.75 & 48.46 & 30.34 \\
    SLP~\cite{liu2022skimming} & 64.35 & 40.43 & 52.89 & 32.04 \\
    D-TSG~\cite{liu2022reducing} & 65.05 & 42.77 & 54.29 & 33.64 \\
    SSRN~\cite{zhu2023rethinking} & 65.59 & 42.65 & 54.49 & 33.15 \\
    \midrule
    Ours & \textbf{68.47} & \textbf{44.92} & \textbf{58.18} & \textbf{37.64} \\
    \bottomrule 
\end{tabular}}
\end{center}
\vspace{-10pt}
\end{table}

\section{Experiments}
\label{sec:exp}

\subsection{Datasets and evaluation protocols}
We evaluate the proposed method on three standard video grounding benchmarks, including the QVHighlights~\cite{lei2021detecting}, Charades-STA~\cite{gao2017tall}, and ActivityNet Captions~\cite{krishna2017dense} datasets.
{\begin{itemize}
    \item{\textbf{QVHighlights} is the recently proposed dataset that supports both video grounding and highlight detection that select representative clips in a given video.
    The dataset contains 10,148 videos with 18,367 moments and 10,310 sentences.
    The dataset also provides annotations of 5-scale saliency scores (from very bad to very good) within the annotated moment for highlight detection.
    Since the official test splits do not have the ground truth, we test on the official validation set.
    }
    \item{\textbf{Charades-STA} includes 16,128 moment-sentence pairs, where the average duration of the full video and annotated moment are 30 and 8.1 seconds long, respectively.
    The official splits provide 12,408 and 3,720 pairs for train and test split.
    }
    \item{\textbf{ActivityNet Captions} contains 15K videos with 72K sentences, where the average duration of the full video and annotated moment are 117.6 and 36.2 seconds, respectively.
    The train, validation\_1, and validation\_2 splits include 37,417, 17,505, and 17,031 moment-sentence pairs.
    Following the previous works~\cite{zeng2020dense,wang2021structured}, we utilize val\_1 for the validation and val\_2 for testing.
    }
\end{itemize}
}

\vspace{-10pt}
\paragraph{Evaluation metrics.}
We adopt Recall1@IoU $m$ following the previous works~\cite{mun2020local,lei2021detecting,chen2021end}; The percentage of top-1 predicted moment having IoU larger than threshold $m$ with the ground truth moment.
We report results with $m=\{0.5, 0.7\}$.
For QVHighlights, we report mean average precision (mAP) with IoU threshold 0.5, 0.75, and the average mAP over IoU thresholds [0.5: 0.05: 0.95].
Although highlight detection is not the task of our interest, we report the results using mAP and HIT@1 for QVHighlights.

\subsection{Implementation details}

\paragraph{Feature representations.}
    {For the QVHighlights dataset, we leverage the pre-trained SlowFast~\cite{feichtenhofer2019slowfast} and CLIP~\cite{radford2021learning} features to extract the video features, following~\cite{lei2021detecting,liu2022umt,xu2023mh,moon2023query} for fair comparisons.
    The features are pre-extracted every 2 seconds.
    For the Charades-STA and ActivityNet Captions datasets, we extract the video features from the most commonly used pre-trained I3D~\cite{carreira2017quo} and C3D~\cite{tran2015learning} backbones, respectively.
    Each feature vector captures 16 consecutive frames with 50\% overlap.
    We uniformly sample 200 feature vectors from each video for ActivityNet Captions dataset.
    For the sentence features, we leverage the token-level CLIP text features.
    }

\vspace{-5pt}
    \paragraph{Training settings.}
    We set the number of layers in the transformer encoder and decoder as $T=3$.
    Following~\cite{lei2021detecting}, {the balancing parameters for the total loss function} are set to $\lambda_{l_1}=10, \lambda_\text{iou}=1$, $\lambda_c=4$, and $\alpha=0.2$.
    $\lambda_\text{sal}$ is set as 1 for QVHighlights, 4 for Charades-STA, and ActivityNet Captions, respectively.
    For all the models, we set the hidden dimensions to 256 and the number of attention heads to 8.
    We train all the models with batch size 32 for 200 epochs using AdamW~\cite{loshchilov2017decoupled} with weight decay 1e-4.
    The initial learning rate is set to 1e-4 for QVHighlights, and 2e-4 for Charades-STA and ActivityNet Captions, respectively.
    All the experiments are implemented with Pytorch v1.12.1 with a single NVIDIA RTX A6000 GPU.

\begin{table}[!t]
\caption{Component ablation results for the proposed method on QVHighlights val split.}
\begin{center} \label{tab:component_ablation}
\resizebox{0.99\linewidth}{!}{
\setlength{\tabcolsep}{6pt}
    \begin{tabular}{
    >{\centering}m{0.16\linewidth}>{\centering}m{0.16\linewidth}
    >{\centering}m{0.16\linewidth}|>{\centering}m{0.16\linewidth}
    >{\centering}m{0.16\linewidth}>{\centering}m{0.16\linewidth}
    }
    \toprule\vspace{-1em}
    \makecell{Event\\reasoning}\vspace{-0.5em} & \vspace{-1em}\makecell{GF\\trans. layer}\vspace{-0.5em} & $\mathcal{L}_\text{event}$ & R1@0.5 & R1@0.7 & mAP \tabularnewline
    \midrule
     & & & 55.10$_{\pm1.4}$ & 40.03$_{\pm1.0}$ & 35.02$_{\pm0.9}$ \tabularnewline
    \cmark & & & 57.71$_{\pm1.4}$ & 42.32$_{\pm0.9}$ & 37.42$_{\pm0.7}$ \tabularnewline
    \cmark & & \cmark & 59.10$_{\pm1.3}$ & 43.32$_{\pm0.9}$ & 38.98$_{\pm0.6}$ \tabularnewline
    \rowcolor{Gray} \cmark & \cmark & \cmark & 61.36$_{\pm1.2}$ & 45.79$_{\pm0.7}$ & 41.74$_{\pm0.7}$ \tabularnewline
    \bottomrule 
\end{tabular}}
\end{center}
\vspace{-5pt}
\end{table}

\begin{figure}[!t]
    \centering
    \captionsetup[subfigure]{oneside,margin={0.5cm,0cm}}
    \begin{subfigure}{0.49\linewidth}
        \centering
        \includegraphics[width=0.99\linewidth]{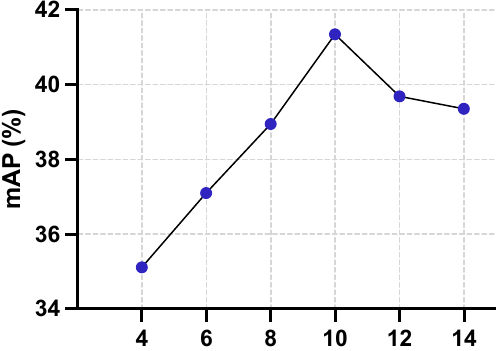}
        \vspace{-8pt}
        \caption{$N\in[4:2:14]$}
        \label{fig:hparamsN}
    \end{subfigure}
    \begin{subfigure}{0.49\linewidth}
        \centering
        \includegraphics[width=0.99\linewidth]{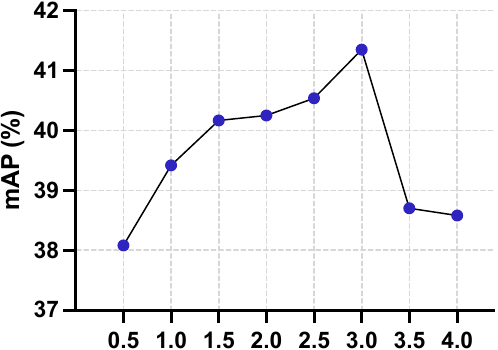}
        \vspace{-8pt}
        \caption{$\lambda_\text{event}\in[0.5:0.5:4]$}
        \label{fig:hparamsE}
    \end{subfigure}
    \caption{Hyper-parameter analysis on QVHighlights val split.}
    \label{fig:hparams}
    \vspace{-5pt}
\end{figure}

\subsection{Comparison with state-of-the-art}
    {In this section, we present the performance comparisons with the state-of-the-art methods~\cite{xiao2021boundary,zeng2020dense,qu2020fine,mun2020local,zhou2021embracing,zhao2021cascaded,liu2020jointly,ding2021support,liu2021context,liu2021progressively,liu2021adaptive,liu2022memory,wang2021structured,liu2022skimming,liu2022reducing,zhu2023rethinking,song2016click,liu2015multi,anne2017localizing,escorcia2019temporal,radford2021learning,lei2020tvr,lei2021detecting,liu2022umt}} to demonstrate the effectiveness of the proposed method.
    Note that we compare the models with the same feature representations for a fair comparison.
    
    \vspace{-10pt}
    \paragraph{Results on QVHighlights.}
    {We provide the performance comparisons between the proposed EaTR and the concurrent DETR-based approaches~\cite{lei2021detecting,liu2022umt,xu2023mh,moon2023query} on QVHighlights~\cite{lei2021detecting}.
    As shown in \tabref{tab:qv}, our EaTR establishes new state-of-the-art performances on several metrics for video grounding, demonstrating the effectiveness of the event-aware dynamic moment queries.
    For highlight detection, our EaTR outperforms Moment-DETR by 1.46\% and 3.05\% in terms of mAP and HIT@1 while showing lower performance than the other approaches~\cite{liu2022umt,xu2023mh,moon2023query}.
    We speculate that the ability to detect highlights mainly depends on learning cross-modal interactions rather than improving the temporal reasoning ability.
    }

    {
    We additionally compare GFLOPs and the number of model parameters to evaluate the computational efficiency.
    While our EaTR requires more parameters than \cite{lei2021detecting,xu2023mh,moon2023query}, we attain an outstanding performance with fewer computations (\ie, GFLOPs) compared to \cite{liu2022umt,moon2023query}.
    }

\vspace{-5pt}
\paragraph{Results on Charades-STA and ActivityNet Captions.}
We report the results evaluated on Charades-STA~\cite{gao2017tall} and ActivityNet Captions~\cite{krishna2017dense} in \tabref{tab:ch_anet}.
The main difference between the previous works~\cite{liu2021adaptive,zeng2020dense,liu2022memory} and our EaTR is the use of hand-crafted components, such as temporal proposals or post-processing steps.
Typically, the proposal-driven approaches~\cite{liu2021context,liu2021adaptive,liu2020jointly} utilize an excessive number of candidate proposals generated with heuristics, while the proposal-free approaches~\cite{zeng2020dense,qu2020fine,liu2022memory} make dense frame-wise predictions to achieve promising results.
These methods require pre-processing steps for defining the candidate proposals or post-processing steps (\eg non-maximum suppression) for reducing the redundant predictions.
Contrary to this, our EaTR utilizes only a set of moment queries (typically less than 10 queries) to predict accurate moments without requiring any hand-crafted components.
Despite this training efficiency, our EaTR outperforms state-of-the-art methods, achieving 2.88\% and 3.69\% performance improvements in terms of R1@0.5 on each dataset.

\begin{table}[!t]
\caption{Component ablation results for the fusion method in the GF transformer layer on QVHighlights val split.}
\begin{center} \label{tab:gf_layer}
\resizebox{0.7\linewidth}{!}{
\setlength{\tabcolsep}{7pt}
\begin{tabular}{lccccc}
    \toprule
    Method & R1@0.5 & R1@0.7 & mAP \\
    \midrule
    Add & 58.68$_{\pm1.4}$ & 43.77$_{\pm0.9}$ & 38.07$_{\pm0.7}$ \\
    Concat & 58.06$_{\pm1.1}$ & 43.10$_{\pm0.7}$ & 38.86$_{\pm0.7}$ \\
    MHCA & 59.74$_{\pm1.2}$ & 44.71$_{\pm0.6}$ & 39.69$_{\pm0.6}$ \\
    \rowcolor{Gray} GF & 61.36$_{\pm1.2}$ & 45.79$_{\pm0.7}$ & 41.74$_{\pm0.7}$ \\
    \bottomrule 
\end{tabular}}
\end{center}
\vspace{-5pt}
\end{table}

\begin{figure}[!t]
    \centering
    \captionsetup[subfigure]{oneside,margin={0.5cm,0cm}}
    \begin{subfigure}{0.49\linewidth}\label{fig:convergea}
        \centering
        \includegraphics[width=0.99\linewidth]{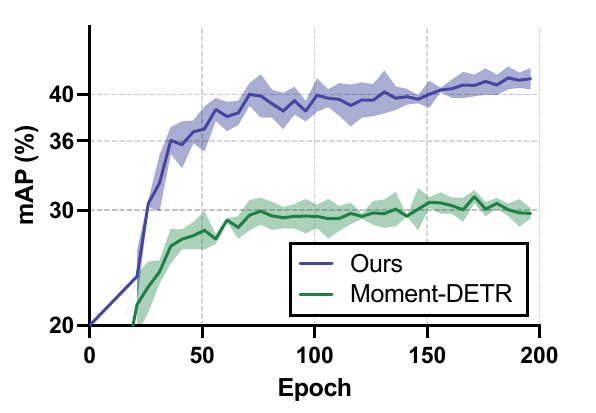}
        \vspace{-10pt}
        \caption{}
    \end{subfigure}
    \begin{subfigure}{0.49\linewidth}\label{fig:convergb}
        \centering
        \includegraphics[width=0.99\linewidth]{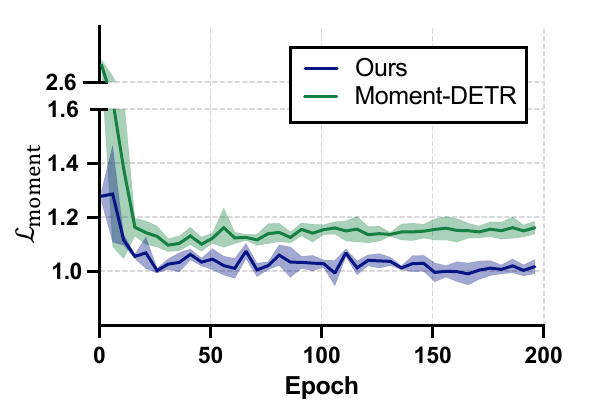}
        \vspace{-10pt}
        \caption{}
    \end{subfigure}
    \caption{Comparison of training convergence of MomentDETR~\cite{lei2021detecting} and Ours on QVhighlight val split. We plot the (a) mAP (\%) curves and (b) $\mathcal{L}_\text{moment}$ curves. We repeat the experiment with 5 different seeds and present the mean performance and range.}
    \label{fig:convergence}\vspace{-5pt}
\end{figure}

\begin{figure*}[!t]
    \centering
    \includegraphics[width=0.99\linewidth]{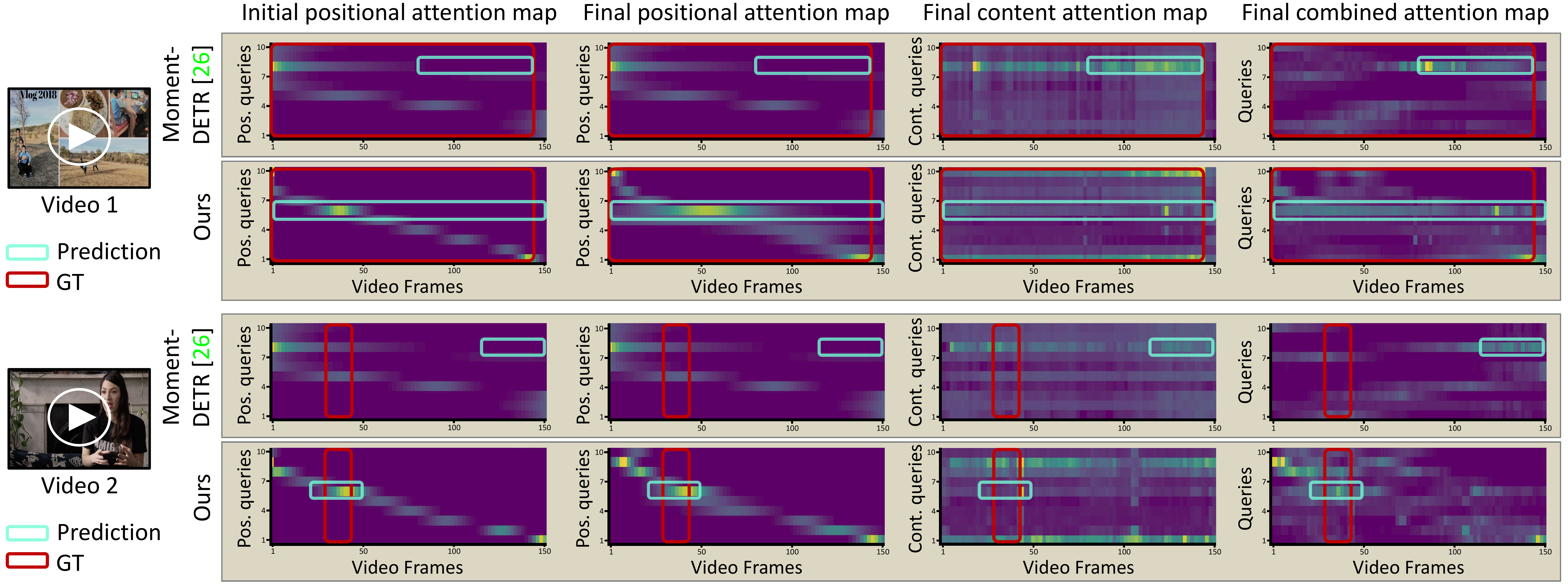}
    \vspace{5pt}
    \caption{Visualization of the positional attention weights, the content attention weights, and the combined attention weights of the cross-attention module of Moment-DETR~\cite{lei2021detecting} and Ours.
    Each attention map is scaled by the corresponding confidence score of each query and sorted in order for clear analysis.
    }
    \label{fig:attention}
    \vspace{-5pt}
\end{figure*}

\subsection{Ablation study and discussion}
    {To investigate the impact corresponding to key components of the proposed method, we conduct ablation studies on the validation set of QVHighlights~\cite{lei2021detecting}.
    In addition, we provide visualization examples to discuss how the dynamic moment queries of the proposed method work.
    The ablation studies for the other two datasets and additional qualitative results are provided in the supplementary material.}

\vspace{-5pt}
\paragraph{Component ablation.}
    We first investigate the effectiveness of each component in our EaTR.
    As shown in \tabref{tab:component_ablation}, we report the impact according to event reasoning, the gated fusion transformer layer, and the event localization loss $\mathcal{L}_\text{event}$.
    Sequentially applying the event reasoning and event localization loss contributes 2.61\% and 4.0\% to performance improvement, and using all components with gated fusion layer improves performance by 6.26\% in terms of R1@0.5.

\vspace{-5pt}
\paragraph{Number of moment queries.}
    {We depict the performance in terms of mAP according to the number of moment queries $N$ in \figref{fig:hparamsN}.
    Since the number of moment queries determines the granularity of the referential search area, the performance is gradually improved as $N$ increases.
    Meanwhile, a large number of $N$ makes it difficult for the model to capture long events, resulting in performance degradation.
    We set $N$ to 10 which represents the best performance.}

\vspace{-5pt}
\paragraph{Effect of $\lambda_\text{event}$.}
    {The balancing parameter $\lambda_\text{event}$ controls the impact of a newly introduced event loss $\mathcal{L}_\text{event}$ in the total training objective.
    To study the sensitivity of $\mathcal{L}_\text{event}$, we report the performance according to $\lambda_\text{event}$ in \figref{fig:hparamsE}.
    The result shows the performance monotonically increases with $1\leq\lambda_\text{event}\leq3$.
    The values of $\lambda_\text{event}$ smaller than 1 or larger than 3 derive the poor performance, validating the impact of the proposed event localization loss.
    }

\vspace{-5pt}
\paragraph{Fusion method in the GF layer.}
    {We conduct an additional ablation study for the fusion method in the GF layer, including the addition, concatenation, and multi-head cross-attention (MHCA) fusions.
    As shown in \tabref{tab:gf_layer}, we observe poor performance with the first two fusion methods.
    While the main goal of the fusion layer is to enhance the sentence-relevant moment queries and suppress the irrelevant ones, the addition and concatenation fusions fail to emphasize the informative moment queries, making the interaction with the video-sentence representations less effective.
    To verify the effectiveness of the gated fusion, we present the performance with the MHCA fusion, which skips \equref{eq:gate} and (\ref{eq:gate_fuse}).
    Although the MHCA provides a slight improvement over simple fusions (\ie, addition and concatenation), it is still insufficient to capture distinctive queries, leading to lower performance than the GF layer.

\vspace{-5pt}
\paragraph{Convergence analysis.}
    {We argued that providing reliable referential search areas accomplish a high training efficiency.
    To validate this, we compare the training convergence of our EaTR and Moment-DETR~\cite{lei2021detecting} which uses input-agnostic moment queries.
    As shown in \figref{fig:convergence}, our EaTR converges much faster and achieves higher performance than Moment-DETR, demonstrating the importance of the precise referential search area when training video grounding models.}

\vspace{-5pt}
\paragraph{Attention visualization.}
    {We validate the impact of our dynamic moment queries on the final prediction.
    As shown in \figref{fig:attention}, we visualize the attention between (first column) the initial positional queries and frame positional embeddings, (second column) the final positional queries and frame positional embeddings, (third column) the final content queries and video-sentence representations, and (last column) the whole moment queries and video-sentence representations with frame positional embeddings.
    To compare the dynamic moment query of our EaTR and the input-agnostic moment query of Moment-DETR, we depict attention maps for two different videos.
    In each attention map, the horizontal and vertical axes represent the frame and query indices, respectively.
    The initial positional queries of Moment-DETR provide a fixed input-agnostic search area regardless of the input video.
    Therefore, the model heavily relies on the quality of the video-sentence interactions, ignoring the highlighted positional attention and leading to the wrong prediction.
    Meanwhile, our EaTR provides different search areas according to the video, making correct predictions with a balanced contribution of the content and positional queries.}

\section{Conclusion and Future Work}
{In this paper, we have introduced a novel Event-aware Video Grounding Transformer, termed EaTR, that performs event and moment reasoning for video grounding.
In event reasoning, we identify the event units comprising a given video with the slot attention mechanism. 
The event units are treated as the initial dynamic moment queries that provide the video-specific referential search areas.
In moment reasoning, we introduce the gated fusion transformer layer to enhance the sentence-relevant moment queries and filter out the irrelevant queries, producing more reliable referential search areas.
The dynamic moment queries interact with the video-sentence representations through the transformer decoder layers, enabling more accurate video grounding over state-of-the-art methods.
Extensive experiments demonstrated the effectiveness and efficiency of the event-aware dynamic moment queries.}

{While we have explored the dynamic moment query based on visual information, consideration of sentence information is still underexplored.
We hope our study will promote the potential of research and provide a foundation for variants of the moment query.
}

\vspace{-10pt}
\paragraph{Acknowledgement.}
This research was supported by the Yonsei Signature Research Cluster Program of 2022 (2022-22-0002) and the KIST Institutional Program (Project No.2E31051-21-203).

{\small
\bibliographystyle{ieee_fullname}
\bibliography{egbib}
}

\clearpage
\newpage

\appendix

\setcounter{table}{0}
\renewcommand{\thetable}{A\arabic{table}}
\setcounter{figure}{0}
\renewcommand{\thefigure}{A\arabic{figure}}

\section*{Appendix}

{In this document, we include supplementary materials for ``Knowing Where to Focus: Event-aware Transformer for Video Grounding''.
We first provide more concrete implementation details of pseudo event timestamps generation (\secref{sec:implementation}), and additional experimental results (\secref{sec:add_exp}), including ablation studies and qualitative results.}

\section{Pseudo event timestamps generation}\label{sec:implementation}
    {We generate the pseudo event-level supervision (\ie, pseudo event timestamps $\hat{\mathbf{P}}$ in Eq. (8)) to learn event reasoning.
    In this section, we describe the details of the pseudo event timestamps generation.
    While pseudo event timestamps generation is highly inspired by the prior work~\cite{kang2022uboco}, which leverages the temporal self-similarity matrix (TSM), we detect pseudo events without any learnable parameters in an unsupervised manner.}

    {Specifically, we first obtain the temporal self-similarity matrix $\mathbf{S}\in\mathbb{R}^{L_v\times L_v}$ by computing cosine similarity between video representations $\mathbf{h}_v$.
    Similar to \cite{kang2022uboco}, we define the contrastive kernel $\mathbf{Z}\in \mathbb{R}^{z\times z}$ with the kernel size $z=5$ as follows:
    \begin{equation}
    \mathbf{Z}=
    \begin{bmatrix}
        1 & 1 & 0 & -1 & -1   \\
        1 & 1 & 0 & -1 & -1   \\
        0 & 0 & 0 & 0 & 0    \\
        -1 & -1 & 0 & 1 & 1   \\
        -1 & -1 & 0 & 1 & 1
    \end{bmatrix}
    \end{equation}
    Since the kernel is designed to imitate the boundary pattern in the TSM~\cite{kang2022uboco}, we can obtain the boundary scores $\mathbf{b}\in\mathbb{R}^{L_v}$ by applying the convolution to the diagonal elements of the TSM.
    With the boundary scores $\mathbf{b}$, we remove the scores that are lower than the average boundary score $\bar{\mathbf{b}}$ and apply a sliding max filter with a size of 3 to filter out the consecutively distributed scores.
    The remaining indices are assumed to be the event boundary, and we define the pseudo event timestamps $\hat{\mathbf{P}}$ as the center coordinate and duration between each boundary index.}

\section{Additional Experiments}
\label{sec:add_exp}
    In this section, we present additional component analysis on QVHighlights~\cite{lei2021detecting} (\secref{sec:component_analysis}), ablation studies on Charades-STA~\cite{gao2017tall} and ActivityNet Captions~\cite{krishna2017dense} (\secref{sec:ablation}), and qualitative results for video grounding (\secref{sec:qualitative}).

\subsection{Additional component analysis}
\label{sec:component_analysis}
    We provide additional component analysis according to the choice of attention mechanism for event reasoning, the number of iterations $K$ in slot attention, the number of transformer layers and qualitative analysis for the gated fusion transformer layer.

\begin{table}[t]
\caption{Choice of attention mechanism for event reasoning on QVHighlights val split.
}
\begin{center} \label{tab:alternative_slot}
\resizebox{0.99\linewidth}{!}{
\begin{tabular}{lccccc}
    \toprule
    Methods & R1@0.5 & R1@0.7 & mAP & GFLOPs & Params \\
    \midrule
    Cross-attention & 57.35$_{\pm1.4}$ & 41.55$_{\pm1.2}$ & 37.00$_{\pm1.0}$ & 0.49 & 10.1M \\
    \rowcolor{Gray} Slot attention & 61.36$_{\pm1.2}$ & 45.79$_{\pm0.7}$ & 41.74$_{\pm0.7}$ & 0.47 & 9.0M \\
\bottomrule 
\end{tabular}}
\end{center}\vspace{-5pt}
\end{table}

\begin{table}[t]
\caption{Performance with respect to the different number of iterations for slot attention on QVHighlights val split.
}
\begin{center} \label{tab:K}
\resizebox{0.75\linewidth}{!}{
\setlength{\tabcolsep}{6pt}
\begin{tabular}{ccccc}
    \toprule
    \textit{K} & R1@0.5 & R1@0.7 & mAP & GFLOPs \\
    \midrule
    1 & 57.16$_{\pm1.6}$ & 41.35$_{\pm0.9}$ & 37.92$_{\pm0.7}$ & 0.467 \\
    2 & 58.26$_{\pm1.2}$ & 43.29$_{\pm0.6}$ & 38.72$_{\pm0.8}$ & 0.469 \\
    \rowcolor{Gray} 3 & 61.36$_{\pm1.2}$ & 45.79$_{\pm0.7}$ & 41.74$_{\pm0.7}$ & 0.472 \\
    4 & 60.45$_{\pm1.3}$ & 44.00$_{\pm0.7}$ & 39.48$_{\pm0.6}$ & 0.474 \\
    5 & 59.16$_{\pm1.2}$ & 43.35$_{\pm0.6}$ & 38.96$_{\pm0.6}$ & 0.476 \\
\bottomrule 
\end{tabular}}
\end{center}\vspace{-5pt}
\end{table}

\vspace{-5pt}
\paragraph{Slot attention vs. cross-attention.}
    {While we use the slot attention mechanism for event reasoning in the main paper, conventional cross-attention can be used as an alternative.
    The main difference between the slot and cross-attention is the attention normalization axis.
    In the cross-attention, the softmax normalization is applied over the input axis, making the attention values for each slot independent of each other.
    Contrary to this, the normalization along event slot direction as in the slot attention enables slots to compete and exchange information with each other to cover distinctive semantics in a given video.
    As shown in \tabref{tab:alternative_slot}, we can obtain higher performance with the slot attention.
    In addition, the slot attention shows higher computational efficiency than the cross-attention in terms of GFLOPs and the number of parameters by reusing the parameters for every iteration.}

\vspace{-5pt}\paragraph{Iteration $K$ in slot attention.}
    {The number of iterations $K$ in the slot attention determines how much each slot interacts with each other.
    To validate the effectiveness of the number of iterations $K$, we evaluate the performance, as shown in \tabref{tab:K}.
    The comparison between $K=1, 2$ and $3$ shows the larger number of $K$ improves the performance with slightly lower computational efficiency (\ie, GFLOPs).
    Meanwhile, larger values of $K$ than 3 bring performance degradation.
    We speculate that a large number of iterations makes the model converges difficult, as analyzed in~\cite{locatello2020object}.
    We set $K$ to 3, which achieves a reasonable trade-off between training efficiency and performance.}

\begin{table}[t]
\caption{Comparison of models with different number of layers on QVHighlights val split.
\# layers indicate the number of transformer encoder-decoder layers used for the video grounding.
}
\begin{center} \label{tab:num_layers}
\resizebox{0.99\linewidth}{!}{
\begin{tabular}{cccccc}
    \toprule
    \# layers & R1@0.5 & R1@0.7 & mAP & GFLOPs & Params \\
    \midrule
    2 & 60.90$_{\pm1.5}$ & 44.06$_{\pm0.9}$ & 38.91$_{\pm0.7}$ & 0.34 & 6.9M \\
    \rowcolor{Gray} 3 & 61.36$_{\pm1.2}$ & 45.79$_{\pm0.7}$ & 41.74$_{\pm0.7}$ & 0.47 & 9.0M \\
    4 & 61.68$_{\pm1.4}$ & 45.90$_{\pm0.7}$ & 41.78$_{\pm0.8}$ & 0.60 & 11.1M \\
    5 & 61.35$_{\pm1.4}$ & 46.94$_{\pm0.8}$ & 41.80$_{\pm0.6}$ & 0.73 & 13.2M \\
\bottomrule 
\end{tabular}}
\end{center}
\vspace{-5pt}
\end{table}

\vspace{-5pt}\paragraph{Number of layers.}
    {We compare the performance according to the number of layers $T$ in \tabref{tab:num_layers}.}
    {Since a small number of layers (less than 3) insufficiently learn the video-sentence interaction, the result shows poor performance.
    While higher performance can be attained with more layers,  the computational complexity also increases.}
Considering the overall performance and efficiency, we set $T$ to 3.

\begin{table}[!t]
\caption{Component ablation results for the proposed method on Charades-STA test split and ActivityNet Captions val\_2 split.}
\begin{center} \label{tab:component_ch_anet}
\resizebox{1\linewidth}{!}{
    \begin{tabular}{
    >{\centering}m{0.16\linewidth}>{\centering}m{0.16\linewidth}
    >{\centering}m{0.16\linewidth}|>{\centering}m{0.12\linewidth}
    >{\centering}m{0.12\linewidth}>{\centering}m{0.12\linewidth}
    >{\centering}m{0.12\linewidth}
    }
    \toprule
    \multirow{2}{*}{\makecell{Event\\reasoning}} & \multirow{2}{*}{\makecell{GF\\trans. layer}} & \multirow{2}{*}{$\mathcal{L}_\text{event}$} & \multicolumn{2}{c}{Charades-STA} & \multicolumn{2}{c}{ANet Captions} \tabularnewline
    \cmidrule(lr){4-5} \cmidrule(lr){6-7}
    & & & R1@0.5 & R1@0.7 & R1@0.5 & R1@0.7 \tabularnewline
        \midrule
         & & & 66.75 & 42.26 & 53.09 & 31.74 \tabularnewline
        \cmark & & & 66.91 & 42.67 & 54.44 & 33.87 \tabularnewline
        \cmark & & \cmark & 67.24 & 43.85 & 55.09 & 35.21 \tabularnewline
        \rowcolor{Gray} \cmark & \cmark & \cmark & 68.47 & 44.92 & 58.18 & 37.64 \tabularnewline
    \bottomrule 
\end{tabular}}
\end{center}
\vspace{-5pt}
\end{table}

\subsection{Ablation study}
\label{sec:ablation}

We provide ablations on the key components of EaTR and hyper-parameters, including the number of moment queries $N$ and the balancing parameter $\lambda_\text{event}$.

\vspace{-5pt}\paragraph{Component ablation.}
We study the impact of each component in EaTR on Charades-STA~\cite{gao2017tall} and ActivityNet Captions~\cite{krishna2017dense} in \tabref{tab:component_ch_anet}.
Each component introduces consistent improvement on both Charades-STA and ActivityNet Captions, where the full usage of components contributes 2.66\% and 5.9\% gain in terms of R1@0.7, respectively.

\vspace{-5pt}
\paragraph{Number of moment queries.}
We depict the impact of the number of moment queries $N$ on Charades-STA~\cite{gao2017tall} and ActivityNet Captions~\cite{krishna2017dense} in \figref{fig:hparamsN_ch} and \figref{fig:hparamsN_anet}.
For Charades-STA, a small $N$ achieves better results than the large $N$ where the optimal result is obtained with $N=6$.
In contrast, for ActivityNet Captions, the overall tendency is similar to the results of QVHighlights~\cite{lei2021detecting} where the optimal result is obtained with $N=10$.
The main difference between Charades-STA and the other two datasets lies in the granularity of videos:
Charades-STA mostly contains fine-grained videos (\ie, visually {similar} with subtle changes) consisting of few events whereas the other two datasets (\ie, QVHighlights and ActivityNet Captions) contain coarse videos (\ie, visually distinct {with significant} changes) consisting of numerous events.
Due to the difference in the granularity of the video, a small number of $N$ is enough for Charades-STA while a large number of $N$ enables the model to better capture the numerous events in videos for QVHighlights and ActivityNet Captions.
Thus, we set $N=6$ for Charades-STA and $N=10$ for ActivityNet Captions.

\begin{figure}[!t]
    \centering
    \captionsetup[subfigure]{oneside,margin={0.5cm,0cm}}
    \begin{subfigure}{0.49\linewidth}
        \centering
        \includegraphics[width=0.99\linewidth]{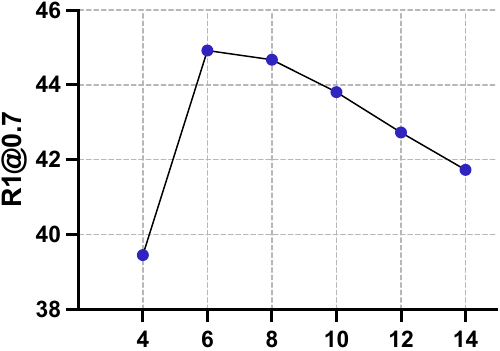}
        \caption{$N\in[4:2:14]$}
        \label{fig:hparamsN_ch}
    \end{subfigure}
    \begin{subfigure}{0.49\linewidth}
        \centering
        \includegraphics[width=0.99\linewidth]{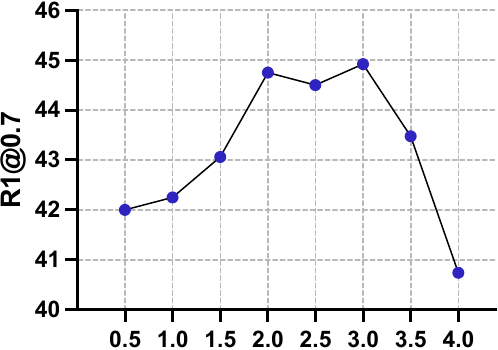}
        \caption{$\lambda_\text{event}\in[0.5:0.5:4]$}
        \label{fig:hparamsE_ch}
    \end{subfigure}
    \caption{Hyper-parameter analysis on Charades-STA test split.}
    \label{fig:hparams_ch}
\end{figure}

\begin{figure}[!t]
    \centering
    \captionsetup[subfigure]{oneside,margin={0.5cm,0cm}}
    \begin{subfigure}{0.49\linewidth}
        \centering
        \includegraphics[width=0.99\linewidth]{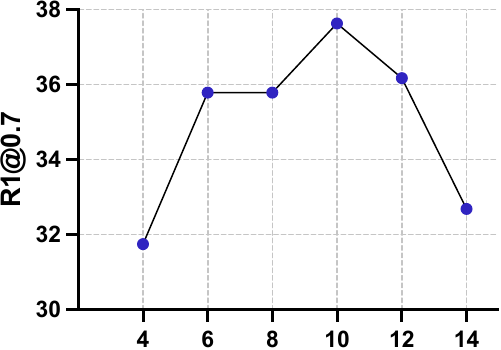}
        \caption{$N\in[4:2:14]$}
        \label{fig:hparamsN_anet}
    \end{subfigure}
    \begin{subfigure}{0.49\linewidth}
        \centering
        \includegraphics[width=0.99\linewidth]{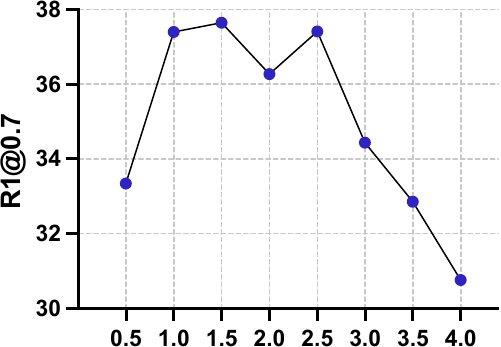}
        \caption{$\lambda_\text{event}\in[0.5:0.5:4]$}
        \label{fig:hparamsE_anet}
    \end{subfigure}
    \caption{Hyper-parameter analysis on ActivityNet Captions val\_2 split.}
    \label{fig:hparams_anet}\vspace{-5pt}
\end{figure}

\begin{figure*}[!t]
    \centering
    \hfill
    \begin{subfigure}{1\linewidth}\centering
        \includegraphics[width=0.85\linewidth]{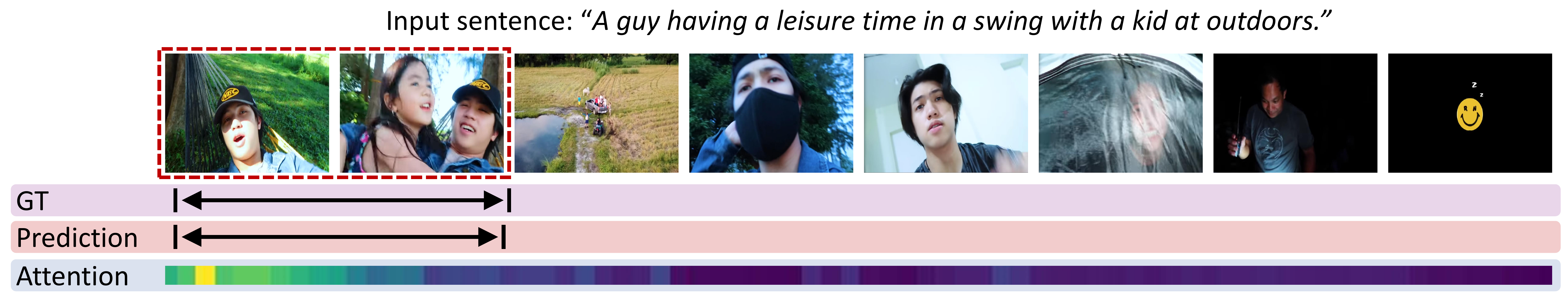}
    \end{subfigure}
    \\
    \centering
    \hfill
    \begin{subfigure}{1\linewidth}\centering
        \includegraphics[width=0.85\linewidth]{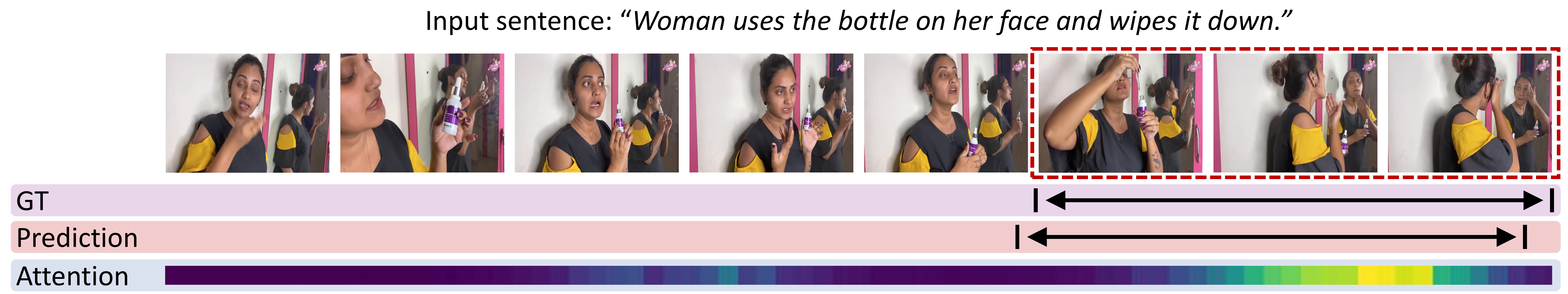}
    \end{subfigure}
    \\
    \centering
    \hfill
    \begin{subfigure}{1\linewidth}\centering
        \includegraphics[width=0.85\linewidth]{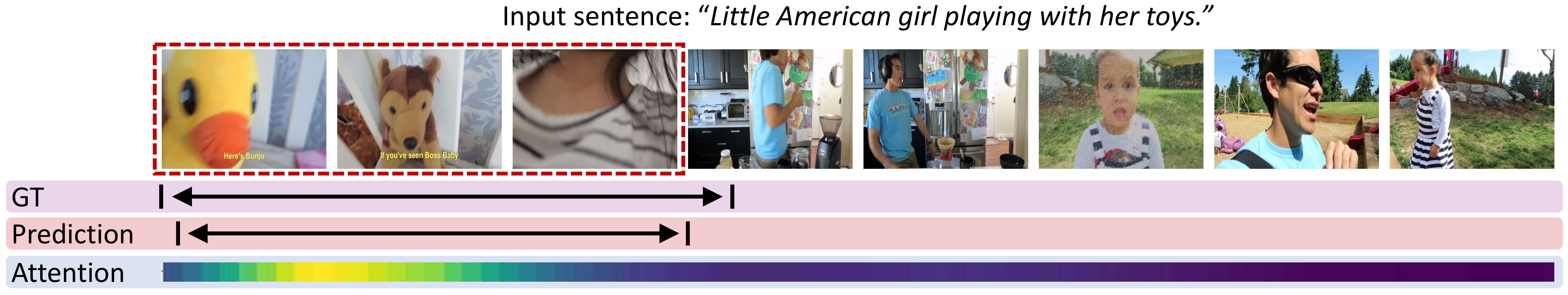}
    \end{subfigure}
    \\
    \centering
    \hfill
    \begin{subfigure}{1\linewidth}\centering
        \includegraphics[width=0.85\linewidth]{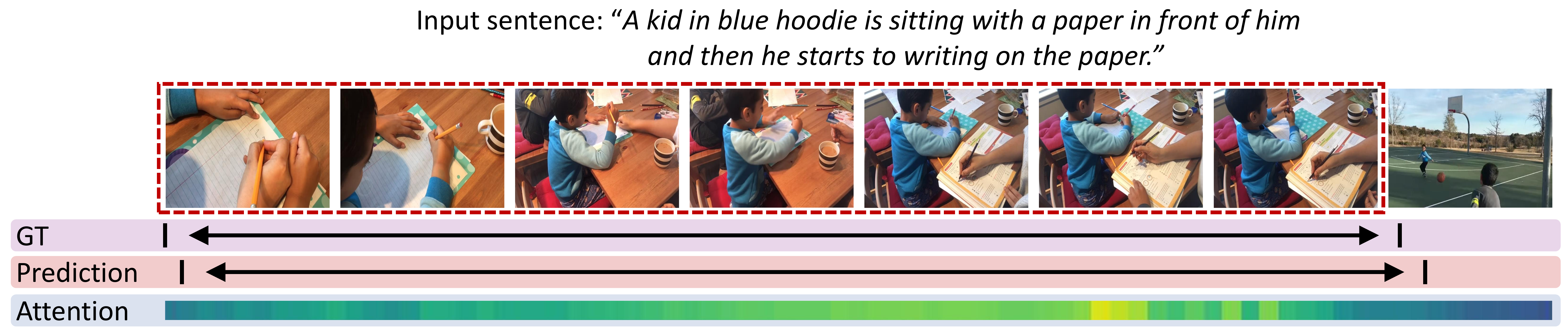}
    \end{subfigure}
    \\
    \caption{Qualitative results of our EaTR on QVHighlights val split.}
    \label{fig:qual_qv}
\end{figure*}

\vspace{-5pt}
\paragraph{Effect of $\lambda_\text{event}$.}
The sensitivity of $\mathcal{L}_\text{event}$ on Charades-STA~\cite{gao2017tall} and ActivityNet Captions~\cite{krishna2017dense} are in \figref{fig:hparamsE_ch} and \figref{fig:hparamsE_anet}.
The event localization loss introduces an improvement with $2\leq\lambda_\text{event}\leq3$ for Charades-STA and with $1\leq\lambda_\text{event}\leq2.5$ for ActivityNet Captions.
The values of $\lambda_\text{event}$ smaller than 1 or larger than 3.5 degrades the performance which is a similar tendency across all three datasets.

\subsection{Qualitative results}
\label{sec:qualitative}
We provide the qualitative results on QVHighlights~\cite{lei2021detecting} and Charades-STA~\cite{gao2017tall} in \figref{fig:qual_qv} and \figref{fig:qual_ch} {to validate the superiority of EaTR on the fine- and coarse-grained videos, respectively}.
We depict the cross-attention weight {from} the last decoder layer computed between the video-sentence representations and the moment queries {that make the final prediction} with the highest confidence score.
Note that we only depicted the attention map corresponding to the video frames for clear analysis.
    {As shown in \figref{fig:qual_qv}, our EaTR correctly localizes the timestamp corresponding to the sentence regardless of the length of the target moment.
    In addition, we provide additional results for a single video labeled with two different sentences in \figref{fig:qual_ch}.
    As shown in the figure, different moment queries are activated according to the given sentence and make the correct final prediction, demonstrating the effectiveness of the event-aware video grounding framework.}

\begin{figure*}[!t]
    \centering
    \hfill
    \begin{subfigure}{1\textwidth}\centering
        \includegraphics[width=0.85\textwidth]{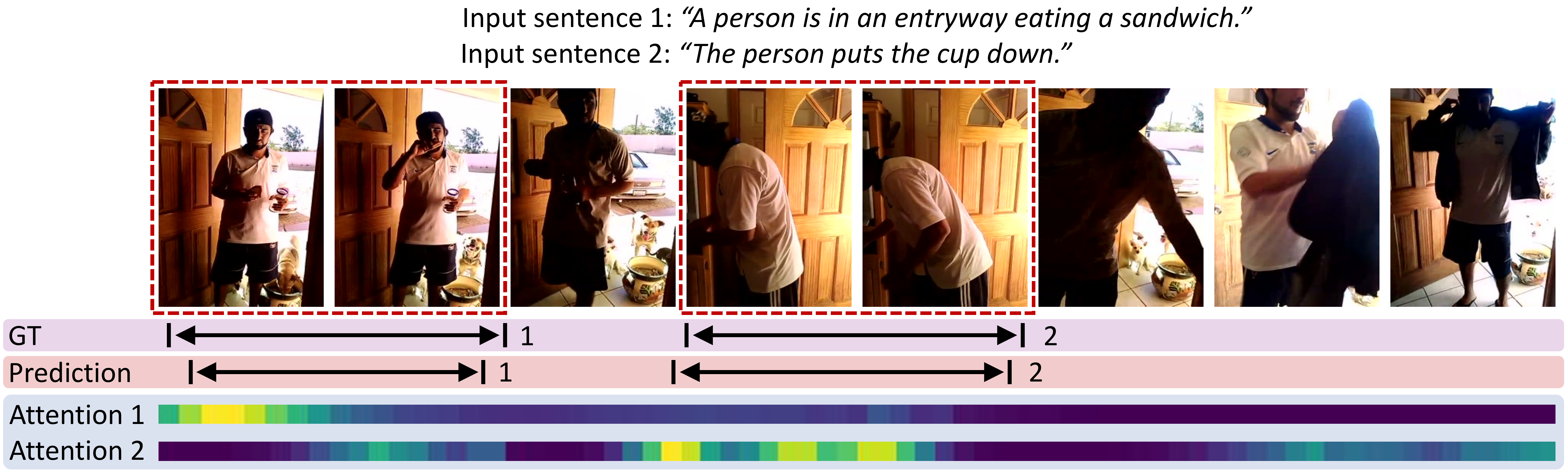}
    \end{subfigure}
    \\
    \centering
    \hfill
    \begin{subfigure}{1\textwidth}\centering
        \includegraphics[width=0.85\textwidth]{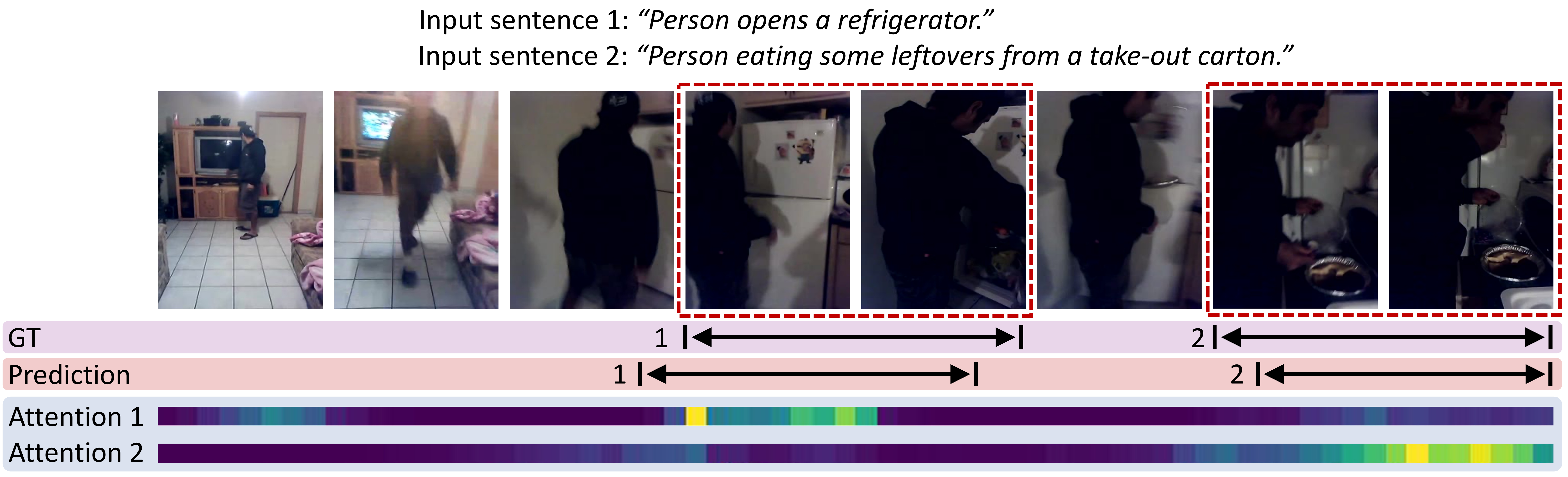}
    \end{subfigure}
    \\
    \centering
    \hfill
    \begin{subfigure}{1\textwidth}\centering
        \includegraphics[width=0.85\textwidth]{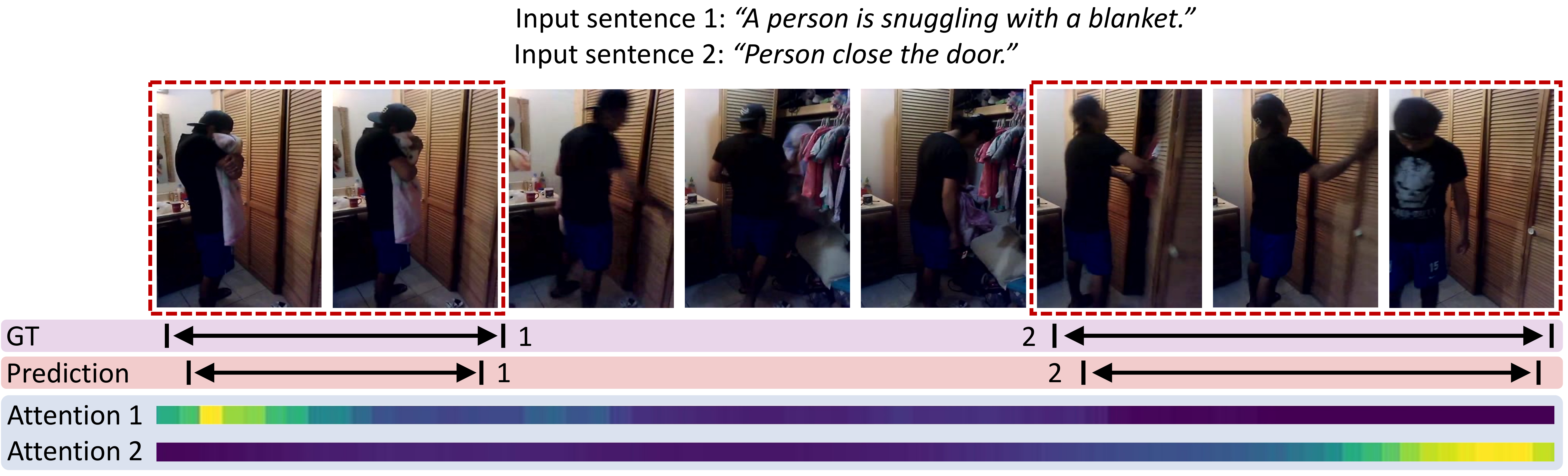}
    \end{subfigure}
    \\
    \caption{Qualitative results of our EaTR on Charades-STA test split.}
    \label{fig:qual_ch}
\end{figure*}

\begin{figure*}[!t]
    \centering
    \hfill
    \begin{subfigure}{1\linewidth}\centering
        \includegraphics[width=0.85\linewidth]{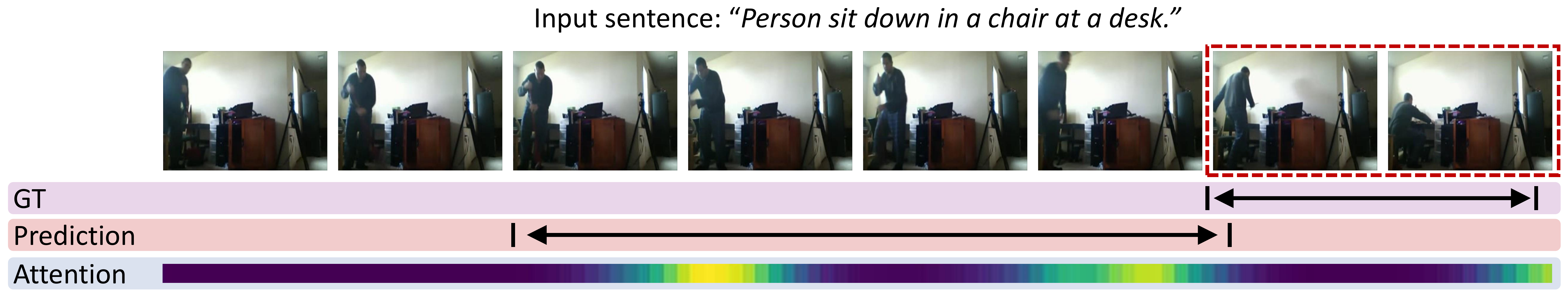}
    \end{subfigure}
    \\
    \centering
    \hfill
    \begin{subfigure}{1\linewidth}\centering
        \includegraphics[width=0.85\linewidth]{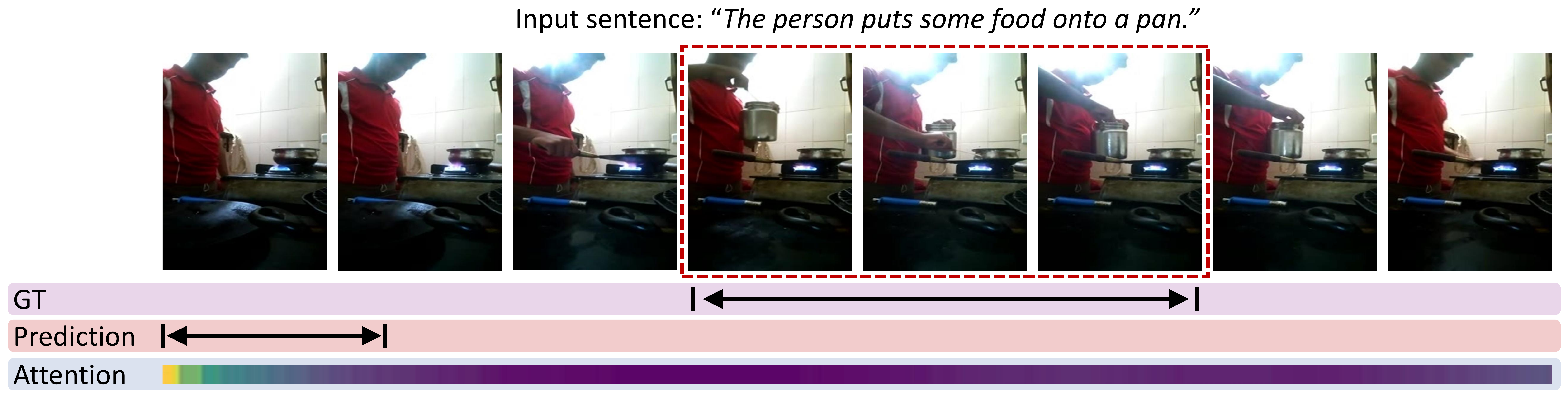}
    \end{subfigure}
    \caption{Failure cases of our EaTR on Charades-STA test split.}
    \label{fig:failure}
\end{figure*}

\vspace{-5pt}
\paragraph{Failure cases.}
Since our EaTR generates the event-aware moment queries based on the visual contents of videos, the model is hard to provide informative referential search area when a video has visually similar frames.
As shown in \figref{fig:failure}, our EaTR fails to localize the given sentence on the fine-grained videos composed of visually similar frames.

\clearpage

\end{document}